\documentclass[10pt,twocolumn,letterpaper]{article}

\usepackage{cvpr}              

\usepackage{graphicx}
\usepackage{amsmath}
\usepackage{amssymb}
\usepackage{booktabs}
\usepackage{comment}
\usepackage{balance}
\usepackage[accsupp]{axessibility}

\usepackage[pagebackref,breaklinks,colorlinks]{hyperref}

\usepackage[capitalize]{cleveref}
\crefname{section}{Sec.}{Secs.}
\Crefname{section}{Section}{Sections}
\Crefname{table}{Table}{Tables}
\crefname{table}{Tab.}{Tabs.}

\def\cvprPaperID{*****} 
\def\confName{CVPR}
\def\confYear{2022}

\begin{document}

\title{An Examination of Bias of Facial Analysis based BMI Prediction Models}

\author{Hera Siddiqui$^1$\textsuperscript{\textsection}, Ajita Rattani$^1$\textsuperscript{\textsection}, Karl Ricanek$^2$, Twyla Hill$^1$\\
$^1$Wichita State University, USA \\
$^2$University of North Carolina Wilmington, USA\\
{\tt\small hxsiddiqui@shockers.wichita.edu, \{ajita.rattani,Twyla.Hill\}@wichita.edu, ricanekk@uncw.edu}}




\twocolumn[{%
\renewcommand\twocolumn[1][]{#1}%
\maketitle
}]
\begingroup\renewcommand\thefootnote{\textsection}
\footnotetext{Equal contribution}
\endgroup

\begin{abstract}
Obesity is one of the most important public health problems that the world is facing today. A recent trend is in the development of intervention tools that predict BMI using facial images for weight monitoring and management to combat obesity.
Most of these studies used BMI annotated facial image datasets that mainly consisted of Caucasian subjects. Research on bias evaluation of face-based gender-, age-classification, and face recognition systems suggest that these technologies perform poorly for women, dark-skinned people, and older adults. The bias of facial analysis-based BMI prediction tools has not been studied until now. This paper evaluates the bias of facial-analysis-based BMI prediction models across Caucasian and African-American Males and Females.
Experimental investigations on the gender, race, and BMI balanced version of the modified MORPH-II dataset suggested that the error rate in BMI prediction was least for Black Males and highest for White Females. Further, the psychology-related facial features correlated with weight suggested that as the BMI increases, the changes in the facial region are more prominent for Black Males and the least for White Females. This is the reason for the least error rate of the facial analysis-based BMI prediction tool for Black Males and highest for White Females. 
\end{abstract}

\section{Introduction}
\label{sec:intro}
Overweight and obesity is a growing epidemic across the world and has been linked to the health disparities associated with the social determinants of health (e.g., age, gender, race, socioeconomic status, sexual orientation, education, health literacy, and access to health care)~\cite{reidpath2002ecological,devaux2013social}.  
World Health Organization (WHO) defines obesity as ``abnormal or excessive fat accumulation that presents a risk to health."\footnote{https://www.who.int/health-topics/obesity}. It was first recognized as a disease in $1948$ by WHO \cite{james2008recognition}. In the United States, it was recognized as a disease by the American Medical Association in $2013$~\cite{pollack2013ama}. 
From $1999–2000$ through $2017–2018$, the age-adjusted prevalence of obesity increased from $30.5\%$ to $42.4\%$, and the prevalence of severe obesity increased from $4.7\%$ to $9.2\%$\footnote{https://www.cdc.gov/obesity/}. By $2030$, several states will have obesity prevalence close to $60\%$, while the lowest states will be approaching $40\%$ and $24.2\%$ will have severe obesity~\cite{ward2019projected}. Obesity is also one of the biggest drivers of preventable chronic diseases and healthcare costs in the United States. Chronic conditions related to obesity include heart disease, stroke, type 2 diabetes, and some cancers, which are the leading causes of preventable death~\cite{national1998clinical, jensen20142013}.

The most common method of indicating overweight and obesity in adults is Body Mass Index~(BMI) which is defined as (body mass in kilograms)/ (body height in meters)$^2$. Overweight individuals have a BMI between $25-30$, and those over $30$ are classified as~\emph{Obese}. 
Besides BMI, other methods of measuring excess body fat are skinfold thickness, waist circumference, underwater weighing, dual-energy x-ray absorptiometry (DXA) etc.~\cite{hu2008measurements}. However, all of these measurements need to be taken by highly trained personnel and are quite expensive.

A number of studies have been proposed for development of machine/deep learning models for (a) obesity prediction ~\cite{degregory2018review, safaei2021systematic, 9627712}, and (b) for understanding the key determinants of obesity~\cite{safaei2021systematic}, for designing intervention strategies. The parametric models, such as Naive Bayes, Support Vector Machines, and Neural Networks, and non-parametric models, such as Decision Trees and K-nearest Neighbour, have been used for obesity prediction. The studies on predictor ranking most commonly used Decision Tree and Gradient Boosting Methods. 

\emph{Self-monitoring} is one of the most important intervention strategies for weight management and lifestyle changes~\cite{robertson2021patterns,laitner2016role}.
A recent trend is in the development of computer vision based self-diagnostic tools for BMI prediction using facial images for weight monitoring and management. For most of these methods, deep convolutional neural network (CNN) such as ResNet or DenseNet is trained on facial images annotated with BMI information for BMI prediction~\cite{Kocabey, 8546159 , jiang2019visual, yousaf2021estimation}. 

Facial analytics has been deployed for various computer vision tasks such as recognition of identities~\cite{albiero2020does}, visual attributes (such as gender, race and age)~\cite{9356331} and deepfake detection~\cite{deepfake,DBLP:journals/corr/abs-2110-01640}.
Recent studies suggest that facial analysis based techniques obtain \textbf{unequal} accuracy rates across demographic variations~\cite{albiero2020does, krishnapriya2020issues,buolamwini2018gender, karkkainen2019fairface, otto2012does, singh2020robustness, singh2021anatomizing}. Specifically, these studies have evaluated bias of face-based gender-, age-classification, and face recognition across gender, race and age-groups.  Most of these studies suggest the bias of the technology for women, dark-skinned people, and older adults~\cite{buolamwini2018gender, karkkainen2019fairface, otto2012does,9356331}. In other words, high error rates have been reported for women, dark-skinned people (like African-American) and older adults.
To date, the bias of facial analysis based BMI prediction tools has \emph{not} been studied systematically. 

The aim of this paper is to \emph{examine the bias of face-based BMI prediction tools} across gender-racial groups. To this front, the bias of deep learning based BMI prediction tools are evaluated across Caucasian and African-American Males and Females using a modified and balanced version of the MORPH-II dataset~\cite{ricanek2006morph}. Worth mentioning, MORPH-II dataset has also been used for bias evaluation of the face-based gender classification and face recognition technology \cite{albiero2020does, qiu2021does}. This is the \emph{only available dataset} with facial images from African-American and Caucasian Males and Females annotated with BMI information.
 
 In summary, the threefold \textbf{contributions} of the paper are as follows:
 
  \begin{itemize}
     
     \item Evaluation of the \textit{bias} of facial-analysis based BMI prediction models across African-American and Caucasian Males and Females.
     
     \item Experimental analysis on the gender, race and BMI \textit{balanced} version of the modified MORPH-II dataset annotated with BMI information.
     
     \item Understanding the \textit{cause of the differential performance} using psychology inspired geometrical facial (PIGF) features related to weight.
     \end{itemize}
     
     This paper is organized as follows: Section \ref{sec:prior} discusses the prior work on facial-analysis-based BMI prediction methods. Section \ref{sec:dataset_protocol} discusses the dataset used and the experimental protocol followed. Experimental results are discussed in section \ref{sec:exp_results}. The psychology-inspired features for understanding the cause of differential performance are discussed in section \ref{sec:psych}. Key findings are listed in section \ref{sec:key_finds}. Discussion is detailed in section \ref{sec:discussion}.

\section {Prior Work on Facial Analysis based BMI Prediction Methods}
\label{sec:prior}

In this section, we will discuss the prior work on BMI prediction from facial images using machine learning and deep learning models. 

Wen and Guo~\cite{wen2013computational} used geometry based features (such as cheekbone to jaw width, width to upper facial height ratio, perimeter to area ratio, and eye size) obtained using Active Shape Model~(ASM)~\cite{milborrow2008locating} with Support Vector Regression~(SVR) for BMI prediction.   
Kocabey et al.~\cite{Kocabey} proposed a facial analysis based BMI prediction method composed of deep feature extraction using VGG-based \cite{simonyan2014very} CNN in combination with Support Vector Regression. 
Dantcheva et al~\cite{8546159} proposed an end-to-end deep learning model obtained by replacing the last fully connected layer of ResNet~\cite{7780459} from $1000$ channels to $1$ channel and using smooth L$1$ loss to cater regression. 

Barr et al.~\cite{barr2018detecting} used a sample of $1412$ predominantly Caucasian young adults to evaluate the algorithm developed by Wen and Guo~\cite{wen2013computational}. The authors compared physically measured BMIs with the BMIs predicted from facial images, and found that $60\%$ of the participants were placed in the correct categories namely, Underweight, Normal, Overweight and Obese using the predicted BMIs.

Jiang et al.~\cite{jiang2019visual} compared three geometry and four deep learning based facial representations for BMI prediction on two datasets, FIW-BMI \cite{jiang2019visual} and Morph-II \cite{ricanek2006morph}. The authors reported that (a) deep-learning models perform better than geometry based methods, (b) dimensionality reduction on deep features from VGG \cite{simonyan2014very} further improves performance, and (c) large head poses degrade the performance of BMI estimation models. In another study~\cite{jiang2020visual}, the authors proposed a two stage approach for BMI estimation from facial images consisting of training face recognition model using centre loss followed by a statistical learning based estimator for BMI prediction. 


Siddiqui et al.~\cite{siddiqui2020ai} evaluated and compared the performance of VGG-19 \cite{simonyan2014very}, ResNet-50 \cite{7780459}, DenseNet-121 \cite{huang2018condensenet}, MobileNet-V2 \cite{howard2017mobilenets}, and lightCNN-29 \cite{wu2018light} for BMI inference from facial images. 
Yousaf et al.~\cite{yousaf2021estimation} used deep features pooled from different face regions (extracted using face semantic segmentation) such as eyes, nose, lips, and eyebrows used for BMI prediction. FaceNet~\cite{schroff2015facenet} and VGGFace~\cite{parkhi2015deep} based CNN models were used for feature extraction. These features from different facial regions were pooled together using Region-aware Global Average Pooling layer. The authors demonstrated an improvement of $22.4\%$ on VIP-attribute, $3.3\%$ on VisualBMI, and $63.09\%$ on Bollywood dataset on using Region-aware Global Average Pooling compared to Global Average Pooling layer.




\begin{table*}[hbt!]
\centering
\caption{Summary of the prior studies on facial analysis based BMI prediction models in terms of dataset, machine/ deep learning models used,
and the obtained error on BMI prediction and obesity classification}
\label{tab:previous-works}
\resizebox{1.0\textwidth}{!}{\renewcommand{\arraystretch}{1.1}{%
\begin{tabular}{|c|c|c|c|l|c|}
\hline 
\textbf{Reference} &
  \textbf{Datasets} & 
  \begin{tabular}[c]{@{}c@{}}\textbf{Feature type}\\ \textbf{(Extraction model)} \end{tabular} &
  \begin{tabular}[c]{@{}c@{}}\textbf{Classification/Regression} \\ \textbf{module} \end{tabular} &
  \textbf{Results} \\ \hline
Wen and Guo\cite{wen2013computational} &
  MORPH-II (Black and Caucasian) & 
  \begin{tabular}[c]{@{}c@{}}PIGF\\(ASM)\end{tabular} &
  SVR &
  \begin{tabular}[c]{@{}l@{}}Overall MAE:\\ {[}3.12– 3.14{]}\end{tabular} \\ \hline
Kocabey et al.\cite{Kocabey} &
  VisualBMI (Caucasian)$^*$ & \begin{tabular}[c]{@{}c@{}}Deep features \\ (VGGFace, VGG)\end{tabular} & \begin{tabular}[c]{@{}c@{}} SVR\end{tabular} &
  \begin{tabular}[c]{@{}l@{}}Pearson correlation:\\ 0.65, 0.47\end{tabular} \\ \hline
Dantcheva et al.\cite{8546159} &
  VIP-attributes (Caucasian)$^*$ & 
  \begin{tabular}[c]{@{}c@{}}Deep features \\ (ResNet-50)\end{tabular}&
  \begin{tabular}[c]{@{}c@{}}end-to-end \end{tabular}&
  Overall MAE: 2.36  \\ \hline
Barr et al.\cite{barr2018detecting} &
  In-house dataset (Caucasian)$^*$ & 
  \begin{tabular}[c]{@{}c@{}}PIGF \\ (ASM)\end{tabular} &
   SVR &
  \begin{tabular}[c]{@{}l@{}}Overall Accuracy:\\ 58.4\%\end{tabular}  \\ \hline
Jiang et al.\cite{jiang2019visual} &
  \begin{tabular}[c]{@{}c@{}}FIW-BMI (Caucasian)$^*$,\\ MORPH-II (Black and Caucasian) \end{tabular} & \begin{tabular}[c]{@{}c@{}}PIGF, PF, PIGF+PF\\ (Openface) \\Deep features \\ (VGGFace, LightCNN-29, \\Centerloss, Arcface) \end{tabular} &
  \begin{tabular}[c]{@{}c@{}} SVR\\ \end{tabular} &
  \begin{tabular}[c]{@{}l@{}} Overall MAE\\ Morph-II: {[}2.30±0.03 - \\3.77±0.08{]} \\ FIW-BMI: {[}3.15±0.07 - \\4.26±0.08] \\ \end{tabular}  \\ \hline
Jiang et al.\cite{jiang2020visual} &
  \begin{tabular}[c]{@{}c@{}}FIW-BMI (Caucasian)$^*$,\\ MORPH-II (Black and Caucasian),\\ VIP-attributes (Caucasian)$^*$ \end{tabular} & 
  \begin{tabular}[c]{@{}c@{}}PIGF, PF, PIGF+PF\\ (Openface) \\Deep features \\ (Centerloss) \end{tabular} &
  \begin{tabular}[c]{@{}c@{}} SVR, PCA-SVR,\\ GPR, CCA, PLS, \\LD-CCA, LD-PLS\end{tabular} &
  \begin{tabular}[c]{@{}l@{}}Best Overall MAE\\ Morph-II:\\ (LD-CCA)—2.42\\ VIP attribute: \\ (LD-CCA)—2.23\end{tabular}  \\ \hline
Siddiqui et al.\cite{siddiqui2020ai} &
  \begin{tabular}[c]{@{}c@{}}VisualBMI (Caucasian)$^*$,\\ VIP attributes (Caucasian)$^*$,\\ Bollywood dataset (Indian)\end{tabular} & 
  \begin{tabular}[c]{@{}c@{}}Deep features \\ (ResNet-50, LightCNN-29\\ MobileNet-V2, VGG-19,\\ DenseNet-121) \end{tabular} &
  \begin{tabular}[c]{@{}c@{}}SVR, RR\end{tabular} &
  \begin{tabular}[c]{@{}l@{}}Overall MAE:\\ {[}1.04,6.48{]}\end{tabular} \\ \hline
Yousuf et al.\cite{yousaf2021estimation} &
  \begin{tabular}[c]{@{}c@{}}VisualBMI (Caucasian)$^*$,\\ VIP attributes (Caucasian)$^*$,\\ Bollywood dataset (Indian)\end{tabular} & 
  \begin{tabular}[c]{@{}c@{}}Deep features\\ (VGGFace, FaceNet) \end{tabular}&
   \begin{tabular}[c]{@{}c@{}}Three layer (512, 256, 1) \\regression module\end{tabular}&
  \begin{tabular}[c]{@{}l@{}}Overall MAE: {[}0.32–\\ 5.03{]}\end{tabular} \\ \hline
  \multicolumn{5}{l}{\begin{tabular}[c]{@{}l@{}}$^*$Majority Caucasians with very few samples from other races.\\ Abbreviations: PIGF, Psychology Inspired Geometric Features; ASM, Active Shape Model; PF, Pointer Features; \\SVR, Support Vector Regression; RR, Ridge Regression; PCA, Principal Component Analysis; GPR, Gaussian Process Regression; \\LD, Label Distribution; CCA, Canonical Correlation Analysis; PLS, Partial Least Square Analysis; MAE, Mean Absolute Error\end{tabular}}
\end{tabular}%
}}
\end{table*}

 \cref{tab:previous-works} summarizes the existing studies on BMI prediction along with the datasets used and the results obtained. The following important \emph{observations} could be drawn from the aforementioned existing studies.

\begin{enumerate}
    \item Studies in \cite{Kocabey, 8546159, siddiqui2020ai, yousaf2021estimation} used three facial image datasets; VisualBMI \cite{Kocabey, siddiqui2020ai}, VIP-attributes \cite{8546159, jiang2019visual}\cite{siddiqui2020ai, yousaf2021estimation}, and FIW-BMI \cite{jiang2019visual, jiang2020visual} for training and evaluation of the BMI prediction tools. These three datasets contain facial images mainly from \emph{Caucasian} subjects. Both VisualBMI and FIW-BMI, are \emph{imbalanced} in terms of gender. VisualBMI has $2,438$ male and $1768$ female images whereas FIW-BMI has $5,197$ images from Males and $2733$ from Females. VIP-attributes has facial images from $513$ Males and $513$ Females but mean BMI values lie mainly between $18$ and $30$ (mean of $25.2$ for Males and $20.9$ for Females). 
    
    \item In~\cite{wen2013computational,jiang2019visual,jiang2020visual}, Morph-II dataset was used but the training set was not balanced across race and gender. The aim of these studies was \emph{not to evaluate the bias} of facial analysis based BMI prediction models.
    
    
\end{enumerate}

\section{Dataset and protocol}
\label{sec:dataset_protocol}

In this section, the gender, race and BMI balanced version of the MORPH-II dataset is discussed followed by the experimental protocol.

The MORPH-II dataset \cite{ricanek2006morph} was originally collected to support research in face aging, and has been widely used in that context. It has also been recently used in the study of demographic variation of facial-analysis based gender classification~\cite{rodriguez2017age, castrillon2017descriptors} and user recognition technology \cite{guo2010cross, vangara2019characterizing}. MORPH-II contains mugshot-style images that are mostly frontal pose, neutral expression and are acquired in a controlled lighting condition.

MORPH-II dataset consists of $55,352$ images with $160$ samples belonging to Asian, $42,722$ Black, $1,753$ Hispanic, $57$ Indian, $10,655$ White, and $5$ belonging to other race categories. For this study, only Black and White races ($53,377$ images) from the Morph-II dataset are considered. Out of $53,377$, only $37,626$ images have height and weight information available. The facial region was cropped from these images using Dlib~\cite{king2009dlib} frontal face detector which is based on Histogram of Oriented Gradients along with linear Support Vector Machine. Among them, $113$ images were discarded as dlib was unable to detect faces in them. 

On analyzing the dataset based on BMI $\geq 30$ (Obese) and BMI $<30$ (Non-obese), a high imbalance in the number of images was observed. There were only $169$ images for White Obese Females whereas for Black Obese Males the number was $3,322$. Similar to White Obese Females, White Males and Black Females had only $484$ and $665$ images with BMI $\geq 30$, respectively. The highest number of images ($22,816$) was for Black Males that had BMI $< 30$.

\begin{figure}[hbt!]
    \centering
    \includegraphics[width=0.45\textwidth]{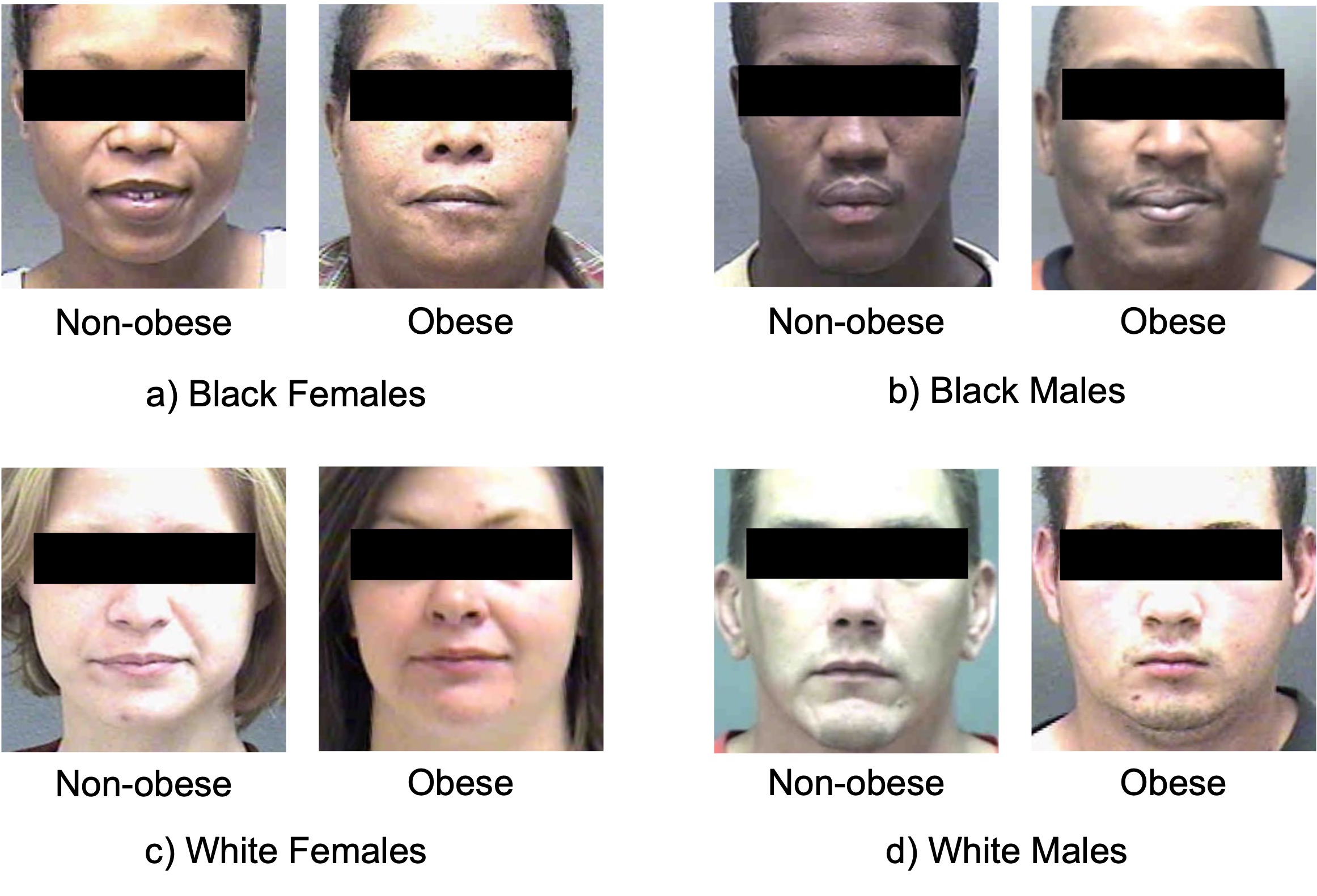}
    \caption{Example facial images from MORPH-II dataset \cite{ricanek2006morph} for Non-obese and Obese Black and White Males and Females.}
    \label{fig:morph1}
\end{figure}

Due to the high imbalance in Morph-II dataset, the high-quality facial images from FIW-BMI~\cite{jiang2019visual, jiang2020visual} and two in-house datasets were used to balance the number of samples for each of the four categories (BF - Black Females, BM - Black Males, WF - White Females, and WM - White Males). The training set of this modified version of the dataset consisted of $9,600$ images with $2,400$ images belonging to each of the four categories i.e. Black Females, Black Males, White Females, and White Males. The test set consists of a total of $3,996$ images. The dataset is \emph{balanced in all aspects i.e., across gender, race, and BMI categories} (Normal, Overweight, and Obese). \cref{fig:morph1} shows examples of facial images from Morph-II dataset for Obese and Non-obese Black and White Males and Females. \\

\noindent \textbf{Deep-learning models:} Five Convolutional Neural Networks (MobileNet-V2~\cite{howard2017mobilenets}, VGG-16~\cite{simonyan2014very}, ResNet-50~\cite{7780459}, EfficientNet-B0~\cite{tan2019efficientnet}, and DenseNet-121~\cite{huang2018condensenet}) pre-trained on ImageNet~\cite{deng2009imagenet} dataset were used in this study.
These models pre-initialized with ImageNet weights were fine-tuned by removing the classification layer and adding two $512$ fully connected layers followed by the output regression layer.
The models were trained on gender, race and BMI balanced training set using an early stopping mechanism with an Adam optimizer and a batch size of $32$. The input to the model were $224\times 224$ aligned face images. Mean Absolute Error~(MAE) was used as the loss function. For the  classification task (normal, over-weight and obese), the output layer consist of three channels and the cross-entropy loss function was used.

Mean Absolute Error (MAE) and accuracy were used for evaluating the performance of the BMI prediction and obesity classification models, respectively. MAE can be defined as the average of the absolute error between predicted BMIs and actual BMIs: $MAE = (\frac{1}{N})\sum_{i=1}^{N}\left | \hat{y_{i}} - y_{i} \right |$, where $\hat{y_{i}}$ and $y_{i}$ are the predicted and actual BMI for i$^{th}$ image and $N$ is the number of images in the test set. Normal, overweight, and obese categories were assigned to an image based on the predicted BMI value. 




\section{Experimental Results}
\label{sec:exp_results}
In this section, we will discuss the results obtained on evaluating the BMI prediction and obesity classification models across gender-racial groups. 

\cref{tab:mae_2} shows the MAE for the five CNN models in BMI prediction for Normal, Overweight, and Obese categories across the four gender-racial groups: Black Females~(BF), Black Males~(BM), White Females~(WF), and White Males~(WM). The overall MAE range across the five CNN models for the four gender-race groups were as follows: BF [$3.55-3.90$], BM [$3.37-3.78$], WF [$4.25-4.64$], and WM [$3.59-4.02$].  
For the three categories (Normal, Overweight, and Obese), the MAE in increasing order were as follows: Normal (WM-$2.60$, BM-$3.07$, BF-$3.19$, WF-$3.51$ ), Overweight (BM-$2.69$, WM-$2.97$, BF-$3.24$, WF-$3.42$), and Obese (BM-$4.81$, BF-$4.85$, WM-$5.65$, WF-$6.39$).

\cref{tab:reg_overall} shows the overall performance of the five models. 
Amongst the five CNN models, ResNet-50 obtained the least MAE of $3.72$. 
Minimum MAE of $3.37$ across all the models was obtained by EfficientNet-B0 for Black Males and maximum MAE of $4.64$ was obtained by VGG-16 for White Females. 


On average, Black Males obtained the \textbf{least} MAE of $3.53$, followed by $3.74$ for White Males. With an MAE of $4.44$, White Females obtained the worst performance. On average, \textbf{Males outperformed females} with an MAE of $3.63$ over $4.10$ obtained by the latter.



Additionally, we also evaluated the classification accuracy of the facial-analysis based methods for Normal (Norm), Overweight (Over) and, Obese (Obese) categories inspired by the study in~\cite{jiang2019visual, jiang2020visual}. \cref{tab:3c_accu} shows the classification accuracy for the three categories (Normal, Overweight, and Obese) across the four gender-racial groups. The worst performing BMI category for all the gender-race groups was the \emph{Overweight} category. It ranged from $30.98\%$ for Black Females to $54.89\%$ for Black Males. The reason being the BMI range of Overweight category lies between those of Normal and Obese category. Hence, there is a higher chance of misclassification of images belonging to lower and higher BMI values in the Overweight range into either of the Normal or Obese categories.

For Black Males, White Females, and White Males the highest accuracy was for the Normal class (BM-$72.79\%$, WF-$69.97\%$, WM-$72.79\%$) followed by the Obese class (BM-$62.52\%$, WF-$59.76\%$, WM-$66.36\%$). For Black Females, the best accuracy was for Obese class ($73.63\%$) followed by Normal class ($68.77\%$). \cref{tab:3c_acc} shows the average classification accuracy values obtained by the five models across the four gender-race groups. The overall accuracy ranged from $56.79\%$ to $61.98\%$. White Males obtained the best overall accuracy of $61.98\%$ amongst the four gender-race category. With only a difference of $0.56$ percentage points from White Males accuracy, Black Males obtained an accuracy of $61.42\%$. 
On average, Males ($61.7\%$) performed better than Females ($57.3\%$). Across models, ResNet-50 obtained the highest overall accuracy of $61.74\%$ with a least standard deviation of $2.08$ across gender-race groups (\cref{tab:3c_acc}). 



In \textbf{summary}, Black Males obtained the best overall MAE of $3.53$ across all four gender-race groups. The best overall classification accuracy was obtained by Black and White Males. With an average MAE and accuracy of $3.63$ and $61.7\%$, Males performed better than Females (MAE-$4.10$, Accuracy-$57.3\%$). Across the five CNN models, ResNet-50 was the best performing model with an average MAE and accuracy of $3.72$ and $61.74\%$ respectively.

\begin{table*}[hbt!]
\centering
\caption{Mean Absolute Error (MAE) of the CNN models in predicting BMI from the facial images for Normal (Norm), Overweight (Over) and Obese categories across four gender-race groups. The least MAE is obtained for Black Males.}
\label{tab:mae_2}
\resizebox{\textwidth}{!}{%
\begin{tabular}{|c|cccc|cccc|cccc|cccc|}
\hline
\textbf{Model} &
  \multicolumn{4}{c|}{\textbf{BF}} &
  \multicolumn{4}{c|}{\textbf{BM}} &
  \multicolumn{4}{c|}{\textbf{WF}} &
  \multicolumn{4}{c|}{\textbf{WM}} \\ \hline
\textbf{} &
  \multicolumn{1}{c|}{\textbf{All}} &
  \multicolumn{1}{c|}{\textbf{Norm}} &
  \multicolumn{1}{c|}{\textbf{Over}} &
  \textbf{Obese} &
  \multicolumn{1}{c|}{\textbf{All}} &
  \multicolumn{1}{c|}{\textbf{Norm}} &
  \multicolumn{1}{c|}{\textbf{Over}} &
  \textbf{Obese} &
  \multicolumn{1}{c|}{\textbf{All}} &
  \multicolumn{1}{c|}{\textbf{Norm}} &
  \multicolumn{1}{c|}{\textbf{Over}} &
  \textbf{Obese} &
  \multicolumn{1}{c|}{\textbf{All}} &
  \multicolumn{1}{c|}{\textbf{Norm}} &
  \multicolumn{1}{c|}{\textbf{Over}} &
  \textbf{Obese} \\ \hline
\textbf{MobileNet-V2}\cite{howard2017mobilenets} &
  \multicolumn{1}{c|}{\textbf{3.88}} &
  \multicolumn{1}{c|}{3.43} &
  \multicolumn{1}{c|}{3.32} &
  4.90 &
  \multicolumn{1}{c|}{\textbf{3.63}} &
  \multicolumn{1}{c|}{2.89} &
  \multicolumn{1}{c|}{2.50} &
  5.51 &
  \multicolumn{1}{c|}{\textbf{4.59}} &
  \multicolumn{1}{c|}{3.49} &
  \multicolumn{1}{c|}{3.87} &
  6.40 &
  \multicolumn{1}{c|}{\textbf{3.92}} &
  \multicolumn{1}{c|}{2.52} &
  \multicolumn{1}{c|}{3.20} &
  6.03 \\ \hline
\textbf{VGG-16\cite{simonyan2014very}} &
  \multicolumn{1}{c|}{\textbf{3.75}} &
  \multicolumn{1}{c|}{2.99} &
  \multicolumn{1}{c|}{3.18} &
  5.07 &
  \multicolumn{1}{c|}{\textbf{3.78}} &
  \multicolumn{1}{c|}{3.36} &
  \multicolumn{1}{c|}{2.96} &
  5.03 &
  \multicolumn{1}{c|}{\textbf{4.64}} &
  \multicolumn{1}{c|}{3.32} &
  \multicolumn{1}{c|}{3.64} &
  6.97 &
  \multicolumn{1}{c|}{\textbf{4.02}} &
  \multicolumn{1}{c|}{2.72} &
  \multicolumn{1}{c|}{3.08} &
  6.25 \\ \hline
\textbf{ResNet-50}\cite{7780459} &
  \multicolumn{1}{c|}{\textbf{3.55}} &
  \multicolumn{1}{c|}{2.88} &
  \multicolumn{1}{c|}{2.92} &
  4.85 &
  \multicolumn{1}{c|}{\textbf{3.41}} &
  \multicolumn{1}{c|}{3.15} &
  \multicolumn{1}{c|}{2.67} &
  4.40 &
  \multicolumn{1}{c|}{\textbf{4.28}} &
  \multicolumn{1}{c|}{3.56} &
  \multicolumn{1}{c|}{3.40} &
  5.88 &
  \multicolumn{1}{c|}{\textbf{3.63}} &
  \multicolumn{1}{c|}{2.71} &
  \multicolumn{1}{c|}{2.79} &
  5.38 \\ \hline
\textbf{EfficientNet-B0}\cite{tan2019efficientnet} &
  \multicolumn{1}{c|}{\textbf{3.75}} &
  \multicolumn{1}{c|}{3.21} &
  \multicolumn{1}{c|}{3.35} &
  4.68 &
  \multicolumn{1}{c|}{\textbf{3.37}} &
  \multicolumn{1}{c|}{2.97} &
  \multicolumn{1}{c|}{2.81} &
  4.34 &
  \multicolumn{1}{c|}{\textbf{4.44}} &
  \multicolumn{1}{c|}{3.60} &
  \multicolumn{1}{c|}{3.16} &
  6.55 &
  \multicolumn{1}{c|}{\textbf{3.59}} &
  \multicolumn{1}{c|}{2.51} &
  \multicolumn{1}{c|}{2.82} &
  5.45 \\ \hline
\textbf{DenseNet-121}\cite{huang2018condensenet} &
  \multicolumn{1}{c|}{\textbf{3.90}} &
  \multicolumn{1}{c|}{3.43} &
  \multicolumn{1}{c|}{3.44} &
  4.76 &
  \multicolumn{1}{c|}{\textbf{3.44}} &
  \multicolumn{1}{c|}{2.99} &
  \multicolumn{1}{c|}{2.53} &
  4.80 &
  \multicolumn{1}{c|}{\textbf{4.25}} &
  \multicolumn{1}{c|}{3.56} &
  \multicolumn{1}{c|}{3.05} &
  6.14 &
  \multicolumn{1}{c|}{\textbf{3.54}} &
  \multicolumn{1}{c|}{2.51} &
  \multicolumn{1}{c|}{2.99} &
  5.12 \\ \hline
\textbf{Average} &
  \multicolumn{1}{c|}{\textbf{3.77}} &
  \multicolumn{1}{c|}{3.19} &
  \multicolumn{1}{c|}{3.24} &
  4.85 &
  \multicolumn{1}{c|}{\textbf{3.53}} &
  \multicolumn{1}{c|}{3.07} &
  \multicolumn{1}{c|}{2.69} &
  4.81 &
  \multicolumn{1}{c|}{\textbf{4.44}} &
  \multicolumn{1}{c|}{3.51} &
  \multicolumn{1}{c|}{3.42} &
  6.39 &
  \multicolumn{1}{c|}{\textbf{3.74}} &
  \multicolumn{1}{c|}{2.60} &
  \multicolumn{1}{c|}{2.97} &
  5.65 \\ \hline
\end{tabular}%
}
\end{table*}

\begin{table*}[hbt!]
\centering
\caption{Statistics of the MAE obtained in Table~\ref{tab:mae_2} }
\label{tab:reg_overall}
\begin{tabular}{|c|c|c|c|c|c|c|c|c|}
\hline
\textbf{Model} & \textbf{BF} & \textbf{BM} & \textbf{WF} & \textbf{WM} & \textbf{Min} & \textbf{Max} & \textbf{Avg} & \textbf{SD} \\ \hline
\textbf{MobileNet-V2}\cite{howard2017mobilenets}   & 3.88 & 3.63 & 4.59 & 3.92 & 3.63 & 4.59 & 4.00 & 0.408 \\ \hline
\textbf{VGG-16}\cite{simonyan2014very}          & 3.75 & 3.78 & 4.64 & 4.02 & 3.75 & 4.64 & 4.05 & 0.415 \\ \hline
\textbf{ResNet-50}\cite{7780459}       & 3.55 & 3.41 & 4.28 & 3.63 & 3.41 & 4.28 & 3.72 & 0.386 \\ \hline
\textbf{EfficientNet-B0}\cite{tan2019efficientnet} & 3.75 & 3.37 & 4.44 & 3.59 & 3.37 & 4.44 & 3.79 & 0.461 \\ \hline
\textbf{DenseNet-121}\cite{huang2018condensenet}    & 3.90 & 3.44 & 4.25 & 3.54 & 3.44 & 4.25 & 3.78 & 0.370 \\ \hline
\textbf{Average}        & 3.77 & 3.53 & 4.44 & 3.74 &      &      &      &       \\ \hline
\end{tabular}
\end{table*}

\begin{table*}[hbt!]
\centering
\caption{Accuracy of the CNN models in classification into Normal, Overweight and Obese categories from facial images. }
\label{tab:3c_accu}
\resizebox{\textwidth}{!}{%
\begin{tabular}{|c|cccc|cccc|cccc|cccc|}
\hline
\textbf{Model} &
  \multicolumn{4}{c|}{\textbf{BF}} &
  \multicolumn{4}{c|}{\textbf{BM}} &
  \multicolumn{4}{c|}{\textbf{WF}} &
  \multicolumn{4}{c|}{\textbf{WM}} \\ \hline
 &
  \multicolumn{1}{c|}{\textbf{All}} &
  \multicolumn{1}{c|}{\textbf{Norm}} &
  \multicolumn{1}{c|}{\textbf{Over}} &
  \textbf{Obese} &
  \multicolumn{1}{c|}{\textbf{All}} &
  \multicolumn{1}{c|}{\textbf{Norm}} &
  \multicolumn{1}{c|}{\textbf{Over}} &
  \textbf{Obese} &
  \multicolumn{1}{c|}{\textbf{All}} &
  \multicolumn{1}{c|}{\textbf{Norm}} &
  \multicolumn{1}{c|}{\textbf{Over}} &
  \textbf{Obese} &
  \multicolumn{1}{c|}{\textbf{All}} &
  \multicolumn{1}{c|}{\textbf{Norm}} &
  \multicolumn{1}{c|}{\textbf{Over}} &
  \textbf{Obese} \\ \hline
\textbf{MobileNet-V2}\cite{howard2017mobilenets} &
  \multicolumn{1}{c|}{\textbf{55.15}} &
  \multicolumn{1}{c|}{63.66} &
  \multicolumn{1}{c|}{29.13} &
  72.67 &
  \multicolumn{1}{c|}{\textbf{59.56}} &
  \multicolumn{1}{c|}{75.98} &
  \multicolumn{1}{c|}{50.45} &
  52.25 &
  \multicolumn{1}{c|}{\textbf{55.25}} &
  \multicolumn{1}{c|}{70.57} &
  \multicolumn{1}{c|}{34.23} &
  60.96 &
  \multicolumn{1}{c|}{\textbf{61.06}} &
  \multicolumn{1}{c|}{75.37} &
  \multicolumn{1}{c|}{43.84} &
  63.96 \\ \hline
\textbf{VGG-16}\cite{simonyan2014very} &
  \multicolumn{1}{c|}{\textbf{57.76}} &
  \multicolumn{1}{c|}{72.37} &
  \multicolumn{1}{c|}{33.30} &
  67.56 &
  \multicolumn{1}{c|}{\textbf{58.76}} &
  \multicolumn{1}{c|}{69.97} &
  \multicolumn{1}{c|}{45.04} &
  61.26 &
  \multicolumn{1}{c|}{\textbf{53.55}} &
  \multicolumn{1}{c|}{71.47} &
  \multicolumn{1}{c|}{36.64} &
  60.06 &
  \multicolumn{1}{c|}{\textbf{57.56}} &
  \multicolumn{1}{c|}{68.77} &
  \multicolumn{1}{c|}{43.84} &
  60.06 \\ \hline
\textbf{ResNet-50}\cite{7780459} &
  \multicolumn{1}{c|}{\textbf{60.96}} &
  \multicolumn{1}{c|}{76.27} &
  \multicolumn{1}{c|}{34.83} &
  71.77 &
  \multicolumn{1}{c|}{\textbf{64.26}} &
  \multicolumn{1}{c|}{71.77} &
  \multicolumn{1}{c|}{54.05} &
  66.97 &
  \multicolumn{1}{c|}{\textbf{59.36}} &
  \multicolumn{1}{c|}{69.07} &
  \multicolumn{1}{c|}{47.14} &
  61.86 &
  \multicolumn{1}{c|}{\textbf{62.36}} &
  \multicolumn{1}{c|}{68.47} &
  \multicolumn{1}{c|}{50.15} &
  68.47 \\ \hline
\textbf{EfficientNet-B0}\cite{tan2019efficientnet} &
  \multicolumn{1}{c|}{\textbf{57.56}} &
  \multicolumn{1}{c|}{67.88} &
  \multicolumn{1}{c|}{27.93} &
  76.88 &
  \multicolumn{1}{c|}{\textbf{62.97}} &
  \multicolumn{1}{c|}{74.17} &
  \multicolumn{1}{c|}{74.17} &
  70.27 &
  \multicolumn{1}{c|}{\textbf{57.05}} &
  \multicolumn{1}{c|}{69.36} &
  \multicolumn{1}{c|}{46.54} &
  55.25 &
  \multicolumn{1}{c|}{\textbf{63.96}} &
  \multicolumn{1}{c|}{75.67} &
  \multicolumn{1}{c|}{49.24} &
  66.96 \\ \hline
\textbf{DenseNet-121}\cite{huang2018condensenet} &
  \multicolumn{1}{c|}{\textbf{57.56}} &
  \multicolumn{1}{c|}{63.67} &
  \multicolumn{1}{c|}{29.73} &
  79.28 &
  \multicolumn{1}{c|}{\textbf{61.56}} &
  \multicolumn{1}{c|}{72.07} &
  \multicolumn{1}{c|}{50.75} &
  61.86 &
  \multicolumn{1}{c|}{\textbf{58.76}} &
  \multicolumn{1}{c|}{69.37} &
  \multicolumn{1}{c|}{46.24} &
  60.67 &
  \multicolumn{1}{c|}{\textbf{64.97}} &
  \multicolumn{1}{c|}{75.68} &
  \multicolumn{1}{c|}{46.85} &
  72.37 \\ \hline
\textbf{Average} &
  \multicolumn{1}{c|}{\textbf{57.80}} &
  \multicolumn{1}{c|}{68.77} &
  \multicolumn{1}{c|}{30.98} &
  73.63 &
  \multicolumn{1}{c|}{\textbf{61.42}} &
  \multicolumn{1}{c|}{72.79} &
  \multicolumn{1}{c|}{54.89} &
  62.52 &
  \multicolumn{1}{c|}{\textbf{56.79}} &
  \multicolumn{1}{c|}{69.97} &
  \multicolumn{1}{c|}{42.16} &
  59.76 &
  \multicolumn{1}{c|}{\textbf{61.98}} &
  \multicolumn{1}{c|}{72.79} &
  \multicolumn{1}{c|}{46.78} &
  66.36 \\ \hline
\end{tabular}%
}
\end{table*}

\begin{table*}[hbt!]
\centering
\caption{Statistics of the classification accuracy values obtained in Table~\ref{tab:3c_accu}.}
\label{tab:3c_acc}
\begin{tabular}{|c|c|c|c|c|l|l|l|l|}
\hline
\textbf{Model} &
  \textbf{BF} &
  \textbf{BM} &
  \textbf{WF} &
  \textbf{WM} &
  \multicolumn{1}{c|}{\textbf{Min}} &
  \multicolumn{1}{c|}{\textbf{Max}} &
  \multicolumn{1}{c|}{\textbf{Avg}} &
  \multicolumn{1}{c|}{\textbf{SD}} \\ \hline
\textbf{MobileNet-V2}\cite{howard2017mobilenets}    & 55.15 & 59.56 & 55.25 & 61.06 & 55.15 & 61.06 & 57.76 & 3.01 \\ \hline
\textbf{VGG-16}\cite{simonyan2014very}          & 57.76 & 58.76 & 53.55 & 57.56 & 53.55 & 58.76 & 56.91 & 2.29 \\ \hline
\textbf{ResNet-50}\cite{7780459}       & 60.96 & 64.26 & 59.36 & 62.36 & 59.36 & 64.26 & 61.74 & 2.08 \\ \hline
\textbf{EfficientNet-B0}\cite{tan2019efficientnet} & 57.56 & 62.97 & 57.05 & 63.96 & 57.05 & 63.96 & 60.39 & 3.58 \\ \hline
\textbf{DenseNet-121}\cite{huang2018condensenet}    & 57.56 & 61.56 & 58.76 & 64.97 & 57.56 & 64.97 & 60.71 & 3.29 \\ \hline
\textbf{Average}        & 57.80 & 61.42 & 56.79 & 61.98 &       &       &       &      \\ \hline
\end{tabular}
\end{table*}

\section{Psychology Inspired Geometric Features}
\label{sec:psych}

 \begin{figure*}[hbt!]
    \centering
    \includegraphics[width=1.0\textwidth]{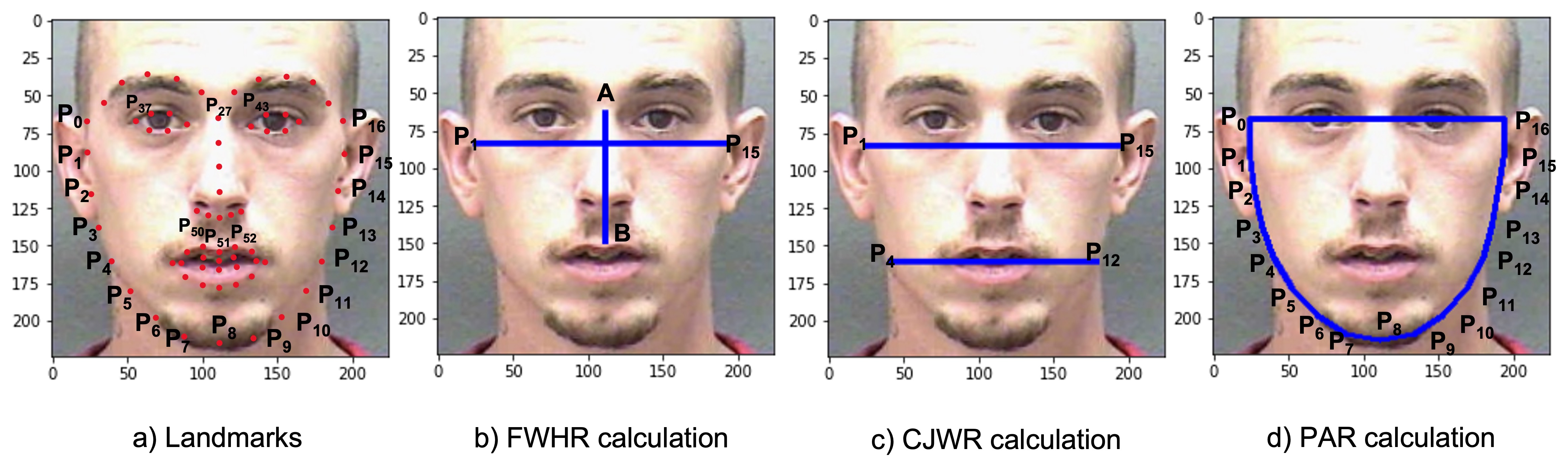}

    \caption{Sample images from Morph-II showing calculation of three Psychology Inspired Geometric Features (PIGF).}
    \label{fig:landmarks}
\end{figure*}

 \begin{figure*}[hbt!]
    \centering
    \includegraphics[width=1.0\textwidth]{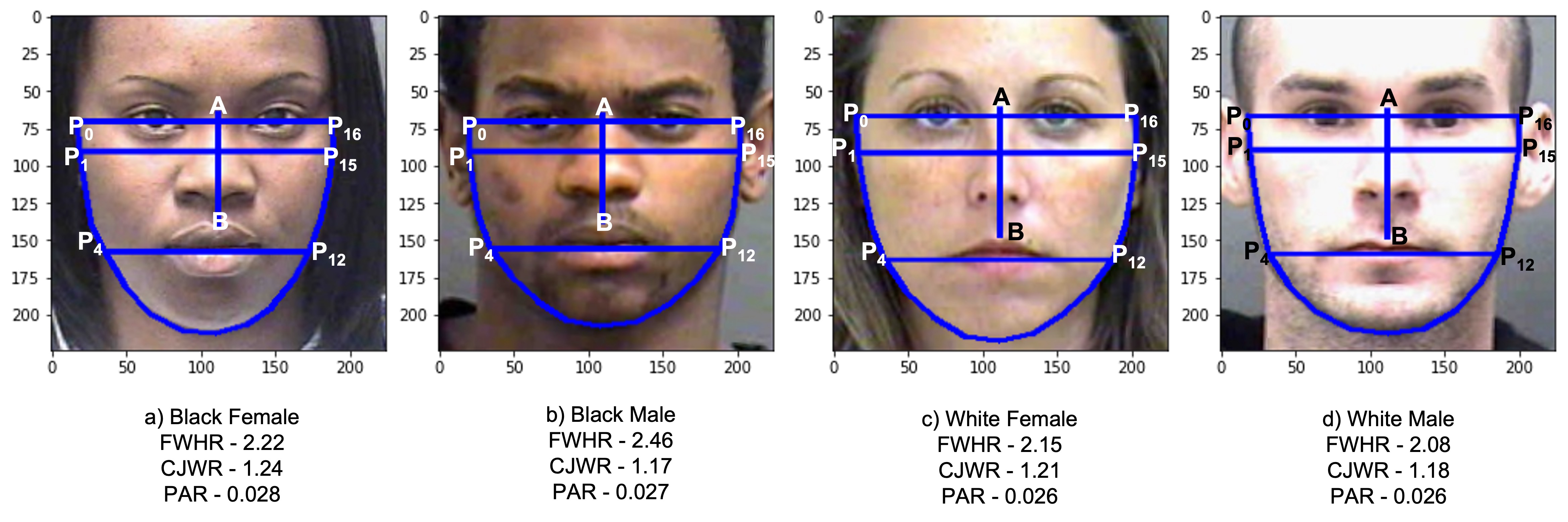}

    \caption{Sample images from Morph-II showing the landmarks used for determining CJWR ($P_1P_{15}/P_4P_{12}$), FWHR ($P_1P_{15}/AB$), and PAR (Perimeter($P_0\ldots P_{16}$)/ Area($P_0 \ldots P_{16}$)) for (a) BF, (b) BM, (c) WF, and (d) WM and the associated values.}
    \label{fig:pigf_all}
\end{figure*}
 

Studies in Psychology~\cite{coetzee2009facial, coetzee2010deciphering} have analyzed correlations between various facial features and weight or BMI. These studies suggested facial features such as cheek to jaw width ratio, width to height ratio, and perimeter to area ratio to be significantly related to BMI or weight of the individual.
Accordingly, \emph{we used these features for understanding the cause of the differential performance of BMI prediction tools} across all the gender-racial groups.

  
 For this set of experiments, frontal images from the test set that had a neutral expression were chosen. For each of the gender-race category, $260$ images were selected (e.g., $130$ Non-obese Black Females and $130$ Obese Black Females). For each of these images, 
pre-trained facial landmark detector from dlib library \cite{king2009dlib}  was used to estimate $68$ landmark points that map to the facial structure. \cref{fig:landmarks}a shows the $68$ landmarks extracted using dlib. Using these landmarks, three facial features namely, facial width to height ratio (FWHR), cheek to jaw width ratio (CJWR), and perimeter to area ratio (PAR), that have been shown to be correlated with BMI were calculated as follows: 
 
 \begin{itemize}
 
  \item \textbf{Facial width to height ratio (FWHR)} - Facial width to height ratio calculation has many variations in the literature. Weston et al. \cite{weston2007biometric} described it as the ratio of cheekbone width to the upper face height (distance between the nasion and prosthion). Carre and McCormick~\cite{carre2008your} adapted this measurement for 2-D facial photographs (due to difficulty in identifying prosthion and nasion in photographs) and defined upper face height as the distance between the most superior point of the upper lip and the most inferior point of the eyebrow. Coetzee et al. \cite{coetzee2009facial} further modified and used the vertical distance between the most inferior point of the upper eyelid and the most superior point of the upper lip as upper facial height. 
  
  For this study, landmark points $27, 37, 43, 50, 51, 52$ in (\cref{fig:landmarks}a and \cref{fig:landmarks}b)  were used for the calculation of upper face height. Point A in \cref{fig:landmarks}b has x-coordinate same as point $27$ and y-coordinate is the average of y-coordinates of points $37$ and $43$. Similarly, the y-coordinate of B was calculated using points $50$ and $52$ and x-coordinate is same as that of point $51$. Cheekbone width is defined as the horizontal distance between the two most lateral facial points \cite{coetzee2010deciphering} \cite{carre2008your}. For cheekbone width calculation, points $1$ and $15$ were used. Therefore, FWHR is the ratio $P_1P_{15}/AB$ in \cref{fig:landmarks}b. \emph{A larger FWHR indicates a wider and squarer face}.
  
 \item \textbf{Cheek to jaw width ratio (CJWR)} - is the ratio of cheek width to jaw width. Cheek width is calculated using points $1$ and $15$ (\cref{fig:landmarks}c) and jaw width (width of the face at the mouth) using points $4$ and $12$. A smaller CJWR ratio i.e. smaller difference between cheekbone width to jaw width indicates a \emph{squarer and wider face}. 

  \item \textbf{Perimeter to area ratio (PAR)} - PAR is defined as the ratio of the perimeter of the lower half of the face to the area of the lower half. In this study, PAR was calculated using points $0$ to $16$. Therefore, it is the ratio of the perimeter of polygon running through these points to the area (\cref{fig:landmarks}d). Smaller PAR signifies a \emph{rounder lower face (also indicates wider and squarer face)}. 
 \end{itemize}
 
\cref{fig:pigf_all} shows sample images from Morph-II dataset with corresponding FWHR, CJWR, and PAR values for the four gender-race categories. Using the calculated FWHR, CJWR, and PAR values, we tested the three \textbf{hypotheses} from~\cite{coetzee2010deciphering}: (a) perimeter-to-area ratio is inversely related to BMI, (b) cheek-to-jaw-width ratio is inversely related to BMI, and (c) facial width-to-height ratio is positively related to BMI. All the three measures were found to be significantly correlated to BMI (\cref{tab:corr}). Comparatively, \emph{the correlations were strongest for Black Males and weakest for White Females}.

\begin{table}[hbt!]
\centering
\caption{Correlations between BMI and the three (FWHR, CJWR, PAR) Psychology Inspired Geometric Features.}
\label{tab:corr}
\begin{tabular}{|c|c|c|c|c|}
\hline
              & \textbf{BF} & \textbf{BM} & \textbf{WF} & \textbf{WM} \\ \hline
\textbf{FWHR} & 0.481$^{**}$     & 0.675$^{**}$      & 0.359$^{**}$      & 0.508$^{**}$      \\ \hline
\textbf{CJWR} & -0.338$^{**}$     & -0.583$^{**}$     & -0.177$^{**}$     & -0.409$^{**}$     \\ \hline
\textbf{PAR}  & -0.492$^{**}$     & -0.549$^{**}$     & -0.293$^{**}$     & -0.459$^{**}$     \\ \hline
\multicolumn{5}{l}{$^{**}$ Correlation is significant at the 0.01 level.}
\end{tabular}
\end{table}

\begin{table*}[hbt!]
\centering
\caption{Percentage increase or decrease in FWHR, CJWR, and PAR from Non-obese to Obese category for BF, BM, WF, and WM.}
\label{tab:support1}
\begin{tabular}{|c|ccc|ccc|ccc|}
\hline
\textbf{} &
  \multicolumn{3}{c|}{\textbf{Non Obese}} &
  \multicolumn{3}{c|}{\textbf{Obese}} &
  \multicolumn{3}{c|}{\textbf{Percentage increase or decrease}} \\ \hline
\textbf{} &
  \multicolumn{1}{c|}{\textbf{FWHR}} &
  \multicolumn{1}{c|}{\textbf{CJWR}} &
  \textbf{PAR} &
  \multicolumn{1}{c|}{\textbf{FWHR}} &
  \multicolumn{1}{c|}{\textbf{CJWR}} &
  \textbf{PAR} &
  \multicolumn{1}{c|}{\textbf{\begin{tabular}[c]{@{}c@{}}FWHR \\    (\%increase)\end{tabular}}} &
  \multicolumn{1}{c|}{\textbf{\begin{tabular}[c]{@{}c@{}}CJWR   \\ (\%decrease)\end{tabular}}} &
  \textbf{\begin{tabular}[c]{@{}c@{}}PAR   \\    (\%decrease)\end{tabular}} \\ \hline
\textbf{BF} &
  \multicolumn{1}{c|}{2.17} &
  \multicolumn{1}{c|}{1.22} &
  0.0278 &
  \multicolumn{1}{c|}{2.34} &
  \multicolumn{1}{c|}{1.2} &
  0.0264 &
  \multicolumn{1}{c|}{7.834} &
  \multicolumn{1}{c|}{1.639} &
  5.04 \\ \hline
\textbf{BM} &
  \multicolumn{1}{c|}{2.19} &
  \multicolumn{1}{c|}{1.2} &
  0.0273 &
  \multicolumn{1}{c|}{2.62} &
  \multicolumn{1}{c|}{1.16} &
  0.0258 &
  \multicolumn{1}{c|}{19.635} &
  \multicolumn{1}{c|}{3.333} &
  5.49 \\ \hline
\textbf{WF} &
  \multicolumn{1}{c|}{2.18} &
  \multicolumn{1}{c|}{1.19} &
  0.0276 &
  \multicolumn{1}{c|}{2.32} &
  \multicolumn{1}{c|}{1.18} &
  0.0269 &
  \multicolumn{1}{c|}{6.422} &
  \multicolumn{1}{c|}{0.840} &
  2.54 \\ \hline
\textbf{WM} &
  \multicolumn{1}{c|}{2.13} &
  \multicolumn{1}{c|}{1.19} &
  0.0271 &
  \multicolumn{1}{c|}{2.36} &
  \multicolumn{1}{c|}{1.17} &
  0.0259 &
  \multicolumn{1}{c|}{10.798} &
  \multicolumn{1}{c|}{1.681} &
  4.43 \\ \hline
\end{tabular}
\end{table*}
 
\cref{tab:support1} shows the average FWHR, CJWR, and PAR for Non-obese and Obese categories across gender and race. The highest increase in FWHR ($19.635\%$) was for Black Males from Non-obese to Obese followed by White Males ($10.798\%$). For Females, the percentage increase in FWHR (WF - $6.422\%$, BF - $7.834\%$) was quite less compared to the Males (WM - $10.798\%$ and BM - $19.635\%$). CJWR decrease was again highest for Black Males ($3.333\%$) followed by White Males at $1.681\%$. White Females obtained a slight decrease of $0.840\%$. Black Male and Black Female obtained the highest PAR\% decrease of $5.49\%$ and $5.04\%$ respectively. White Females obtained the lowest PAR percentage decrease ($2.54\%$). 

These results point out that \textit{Obese Black Males have a wider, squarer face, and rounder chin compared to the Non-obese Black Males} over other gender-racial groups. This aids facial analysis tools in accurate BMI prediction for Black Males. In the case of White Females, the percentage increase or decrease in FWHR, CJWR, and PAR is not as high as compared to other gender-racial categories. This indicate that for White Females not much change is evident in the face from Non-obese to Obese category. Thus, obtaining least performance for facial analysis based BMI prediction tools. This is also evident for females in general, which explains the cause of females \textit{under-performing} males for facial analysis based BMI prediction tools.
\section{Key findings}
\label{sec:key_finds}
Following are the important findings and observations from the experiments conducted:
\begin{itemize}
    \item Black Males obtained the least MAE ($3.53$) across all gender-race groups.
    \item Black Males and White Males obtained the highest classification accuracy of about $61.00\%$ for normal, overweight and obese categories.
    \item White Females obtained the worst MAE ($4.44$) as well as accuracy ($56.79$) amongst all the gender-race groups.
    \item Males obtained a lower average MAE and higher accuracy of $3.63$ and $61.7\%$ respectively as compared to Females (MAE-$4.10$, Accuracy-$57.3\%$).
    \item Psychology related features suggested that compared to other gender-racial groups, Obese Black Males have wider, square, and rounder faces compared to Non-obese Black Males. For White Female, not much change is evident in the face from Non-obese to Obese category. This explains the reason for Black Males outperforming and White Females under-performing for BMI prediction tools.
\end{itemize}

\section{Discussion}
\label{sec:discussion}


The aim of this study was to evaluate the bias of face-based BMI prediction models across four gender-racial groups (Black Females, Black Males, White Females, and White Males). 
Experimental results suggested performance differential of facial analysis-based BMI prediction tools. 
However, \emph{in-contrary} to bias analysis of other computer vision systems reporting the least performance for dark-skinned people~\cite{lohr2018facial, buolamwini2018gender,singh2020robustness, singh2021anatomizing}, Black Males obtain the least error rate in BMI prediction from facial images in this study. The psychology-related features suggested that as the BMI increases, the changes in the facial region are more prominent for Black Males than any other gender-race category. 
This \emph{assists the BMI prediction tools based only on facial image analysis} in more accurate prediction for Black Males over other gender-racial groups. In our experiments, Males outperformed females in BMI prediction which is also the general trend reported for other computer vision applications~\cite{lohr2018facial, buolamwini2018gender}.


With the increasing interest in facial-analysis-based self-monitoring tools as an intervention strategy to combat obesity, it becomes vital to examine and mitigate the bias of this technology. 
To the best of our knowledge, MORPH-II is the only dataset with the BMI annotated facial images from African-American and Caucasian subjects.  
To promote further research and development in this area, the path forward would be large-scale BMI annotated facial image dataset collection across demographics. This should be followed by a thorough evaluation of the bias of this technology. Accordingly, methods to mitigate the bias of this technology should be developed to ensure equal access to health care tools and for promoting well-being among all diverse population sub-groups.

\balance
{\small
\bibliographystyle{ieee_fullname}
\bibliography{paper}
}

\end{document}